\documentclass[conference]{IEEEtran}

\usepackage{graphicx,siunitx,tcolorbox,subcaption}
\usepackage{amsmath}
\usepackage{amssymb}
\usepackage{mathrsfs}
\usepackage{lipsum}
\usepackage{mathtools}
\usepackage{cuted,textcomp}

\usepackage[strict]{changepage}
\usepackage{calc}

\usepackage{lipsum}

\begin{document}
%
\title{Predicting the Voltage Distribution for Low Voltage Networks using Deep Learning}


\author{\IEEEauthorblockN{Maizura Mokhtar, Valentin Robu}
\IEEEauthorblockN{David Flynn}
\IEEEauthorblockA{\textit{Heriot-Watt University}\\
Edinburgh, Scotland } 
\and
\IEEEauthorblockN{Ciaran Higgins}
\IEEEauthorblockA{\textit{Derryherk Ltd.} \\
Scotland }
\and
\IEEEauthorblockN{Jim Whyte}
\IEEEauthorblockA{\textit{NotSoAnalytic Ltd.} \\
Scotland }
\and
\IEEEauthorblockN{Caroline Loughran}
\IEEEauthorblockN{Fiona Fulton}
\IEEEauthorblockA{\textit{SP Energy Networks} \\
Scotland }
\thanks{This work is part of the NCEWS and NCEWS2 projects funded by Ofgem (Network Innovation Allowance NIA\_SPEN0016 and NIA\_SPEN\_034) and Innovate UK (B16N12241).}}

\IEEEoverridecommandlockouts
\IEEEpubid{\makebox[\columnwidth]{978-1-5386-8218-0/19/\$31.00 \copyright2019 IEEE \hfill} \hspace{\columnsep}\makebox[\columnwidth]{ }} 

\maketitle

\begin{abstract}
The energy landscape for the Low-Voltage (LV) networks are beginning to change; changes resulted from the increase penetration of renewables and/or the predicted increase of electric vehicles charging at home.
The previously passive `fit-and-forget' approach to LV network management will be inefficient to ensure its effective operations. A more adaptive approach is required that includes the prediction of risk and capacity of the circuits. Many of the proposed methods require full observability of the networks, motivating the installations of smart meters and advance metering infrastructure in many countries. However, the expectation of `perfect data' is unrealistic in operational reality. 
Smart meter (SM) roll-out can have its issues, which may resulted in low-likelihood of full SM coverage for all LV networks. This, together with privacy requirements that limit the availability of high granularity demand power data have resulted in the low uptake of many of the presented methods. To address this issue, Deep Learning Neural Network is proposed to predict the voltage distribution with partial SM coverage. The results show that SM measurements from key locations are sufficient for effective prediction of voltage distribution. 
\end{abstract}


%
\IEEEpeerreviewmaketitle

\section{Introduction}
\label{ami}

One of the key motivators for the smart meters (SM) and advance metering infrastructures (AMI) installations is to increase observability to the Low Voltage (LV) networks. Observability is required for the active management of the LV network. Active management will soon become a necessity  to cope with the predicted changes to the customers' energy behaviour; changes that include the predicted increase in electric vehicles charging at home, electrification of heat and renewables.

Smart meter installation, however in the UK especially, is on a best endeavours bases, with a number of factors that can impact on the reliability of the SM data transfer to the distribution network operators or DNOs. These factors, for example,  the reliability of the communication infrastructure and customer's choice. 
Because of these key reasons, from a DNO perspective, 
there is a high likelihood of incomplete SM coverage. In this paper, the LV networks that are of interest is the 400V and below networks, specifically those providing electricity to domestic properties. We define an LV circuit as the cable topology that connects customers to the same secondary substation that is providing electricity to them. An LV network is defined as a collection of LV circuits in a specific area. 
The lack of full SM coverage for these LV networks may have resulted in the low uptake of many Distribution State System Estimation (DSSE) techniques presented in literature. 

The applicability of many of these DSSE techniques are also affected by customers' privacy needs.
In the UK for example, its energy regulator, the Office of Gas and Electricity Markets or Ofgem, indicates that {\em''household-level electricity data that relates to a period of less than a month is considered to be personal data''} \cite{ofgem_privacy}. Ofgem has enforced strict measures to be in place to ensure that customers privacy needs are maintained. 
Ofgem requires UK DNOs to submit a data privacy plan that indicates  
how they are to manage the personal data, indicating the steps to ensure privacy and data security of the customers are maintained. As a result, high granularity demand power data will come at a 
cost to UK DNOs. 
An example method to meet this requirement is to aggregate the demand power data for $>$1 nearby households on the LV circuit \cite{NIJHUIS201759}. However, to ensure its applicable for DSSE, aggregation should be done for those on the same phase. This can be an issue because information regarding customer phasing is often unavailable.

At present, only the total energy and the maximum demand power per month per SM (customer) are available to SP Energy Networks. This is in advance of their data privacy plans being accepted by Ofgem.
Such restriction however is not applicable to the voltage data per SM. Half hourly average $V_{rms}$ are provided by the SM. Therefore,
the evaluation of techniques for LV network analysis using voltage and minimal to zero demand power data is required to determine what level of personal demand power data is necessary. 
This has motivated the research presented in this paper, to propose a method that can predict how voltage is distributed across an LV circuit using minimal SM data and reduced granularity for the customer demand power data. 
This paper proposes the use of Deep Learning Neural Network (DLNN) to achieve this aim. The ability to predict the voltage distribution is significant for real-time risk identification of the LV circuits, specifically the prediction of voltage constraints violation of the LV circuits.

This paper is divided into 5 sections.  
Section \ref{challenge} discusses the limitations and challenges of existing DSSE techniques presented in literature, and the motivation behind our proposed method.   
The number of visible SM available to the DNO in any LV circuit is an uncontrollable value. This paper therefore proposed the use of DLNN to predict the voltage distribution across the LV circuit with varying degree of SM coverage. This is presented in Section \ref{ANN}. Section \ref{results}
describes the results from our evaluation. Section \ref{concludes} concludes the paper. 

\section{Challenges \& Motivation}
\label{challenge}

Distribution State System Estimation (DSSE) tools are often proposed to predict the voltage distribution across the LV networks. These methods were proposed for 
demand side management (DSM), renewables integration  \cite{Stetz2014}, \cite{Ferreira2013}, or both \cite{Hayes2016}. All of these presented techniques assumed that demand power data  from all customers in the circuit are available at high granularity, half-hourly or less. As indicated in Sec. \ref{ami}, the likelihood of full coverage of SMs is low. Individual demand power data at high interval may not be provided from all customers because of  privacy restrictions.

To overcome this limitations, pseudo-measurements are used \cite{MCKENNA2016445}, \cite{NIJHUIS201759}. 
The key disadvantage of pseudo-measurements is the potential error propagation from  the pseudo-measurement to the output of the DSSE, error which  increases the level of uncertainty of the results \cite{Papadopoulos2012}.
Furthermore, the uncertainty with regards to customers' phase connection also affects the quality of the results. 

In the UK, for example, nearly all of its domestic electricity users are connected to the LV circuit using a single-phase cable with the voltage tolerance between $-6\%$ to $+10\%$ of the referenced voltage 1pu (230V). These individual phases are taken from the three-phase mains cable. As a result, one of the key identifiable challenges for UK LV network management is the missing customer phase information. It is assumed that the LV circuit is a near-balance circuit, however, in reality, this is not often the case. 
Identification of customer phase is an active area of research, with voltage clustering \cite{Deka2018} and energy data correlation \cite{Park2018} are some of the methodologies proposed for customer phase identification. The later technique is more suitable if, for each customer on the circuit, high granularity demand power data at every half hour or less is available. This method will not be applicable if the customer privacy concerns reduces the granularity of the demand power data.  The two algorithms are not applicable 
if there is the high likelihood of incomplete SM coverage in an LV circuit. 

\begin{figure}[!t]
	\centering
	\includegraphics[width=0.47\textwidth]{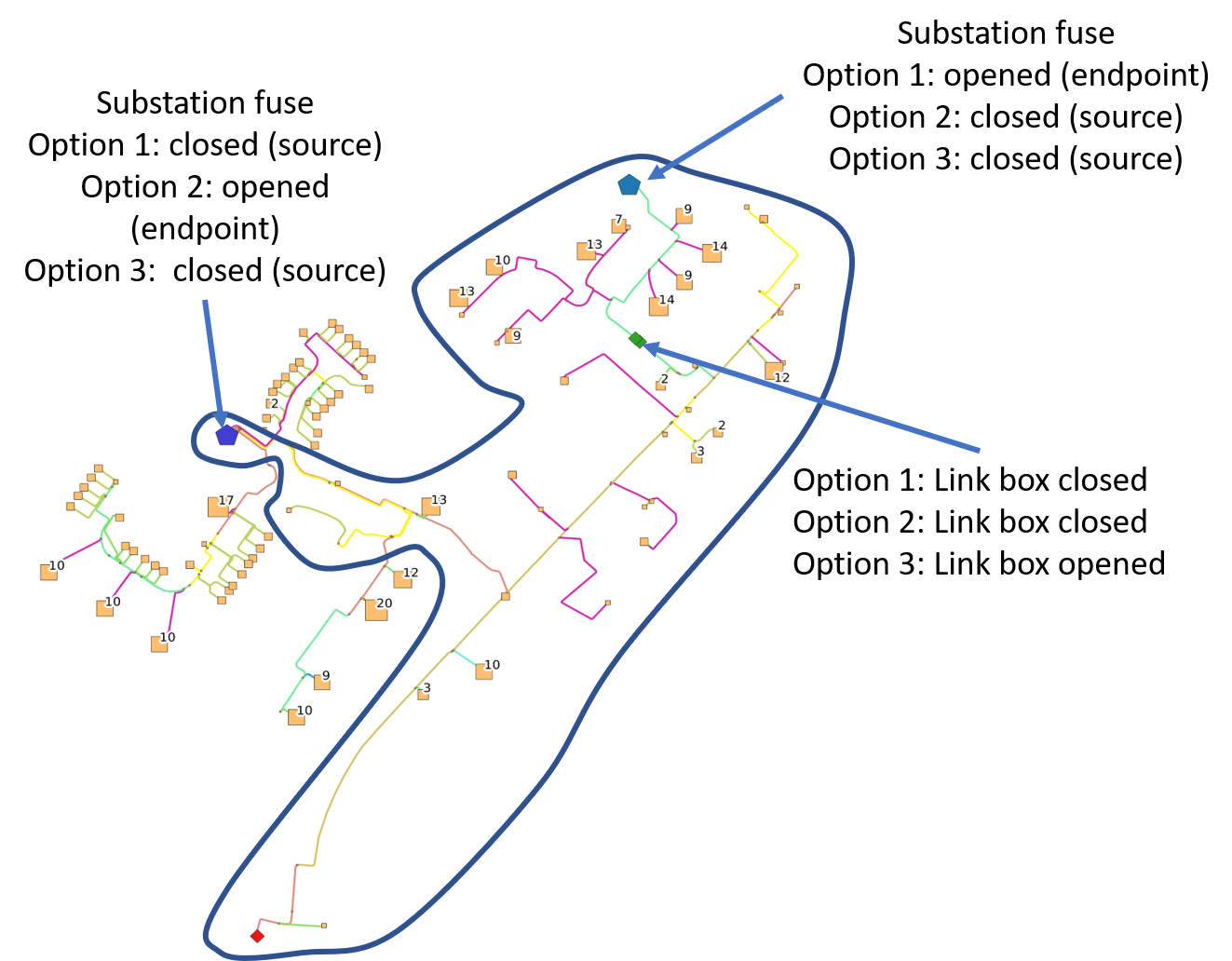}
	\caption{The LV circuit (encircled area) has 3 topology options by changing the state of its fuses and link box.}
	\label{fig:options}
\end{figure}

A new approach is therefore required to predict the voltage distribution using only the available information provided. 
We are also evaluating the  accuracy of prediction for varying degree of observability, by varying the number of SMs on the LV circuit.
The ability to predict 
the voltage distribution across a circuit ahead of time is important because of a number of factors. Voltage prediction is required for effective DSM and renewable import to the network. Furthermore, most urban (and radial) LV circuits in the UK have the flexibility to change its topologies. For example, the LV circuit displayed in Fig. \ref{fig:options}. The circuit encircled has three topology options just by changing the state of its fuses and link box. Each option will have its own risk and capacity. At present, any connectivity improvement initiatives are performed manually when faults or change of requirements are reported. 
Dynamic connectivity improvement strategy or the ability to change the circuit topology based on the predicted risk may soon be required because of 
the potential mismatch between their renewables output and the peaks in energy demand. This strategy is important, especially when DSM are unable to alleviate the risk of constraints violation. 
The ability to predict risks of constraint violations via the prediction of voltage distribution is essential to instigate the need for connectivity improvement. 

\section{Methodology}
\label{ANN}
This paper proposed the use of Deep Learning Neural Network (DLNN) to predict the voltage distribution in an LV circuit. Only the voltage magnitude is predicted, as this value is of interest to DNO for the $\leq$400V LV circuits \cite{Pertl2016}.
Because of the limitations identified in Sec. \ref{ami}, the model aims are: 
\begin{enumerate}
	\item To predict how voltage is distributed across a circuit one-time step ahead $(t+\tau)$, despite the partial SMs coverage; 
	\item To predict the voltage for all customers, including at location without any SMs;
	\item To predict without the need for high granularity demand power data from all customers on the circuit, including pseudo-measurements data. The prediction is performed using only the aggregated demand power data or no demand power data;
	\item To predict without prior identification of customers phase connection.
\end{enumerate}

\subsection{Simulating the voltage distribution}
\label{sec:simulate}
Because of current lack of large-scale SM data, 
simulated SM data are used to evaluate our assumptions. OpenDSS 
is used to simulate the voltage distribution across an LV circuit, using actual LV circuit topologies randomly selected from Scotland Central Belt and the demand power data per household is generated from University of Loughborough CREST model \cite{MCKENNA2016445}. The original CREST model was modified to enable the generation of 1-month demand power data with no renewable generation.

For our analysis, we define where the service cable connects to the property as the customer connection point (CCP). A CCP can connect to a single household or multi-household property. 
A single household property will typically have a single-phase CCP. For a single phase CCP connected to a single household property, the values provided by the simulated SM will be similar to reality. However, simulation of the SM may differ from reality for multi-households properties.

Multi-household property, for example, flats and tenements, can have a maximum of 3 CCPs, one for each phase; with the number of households in a property are equally distributed across the 3 phases, assuming balanced loading. For example, if there are 6 households in a multi-household property, an example of a tenement housing, each single phase 230V cable will be connected to 2 households in the property. 
No lateral cables or cables in the property are provided for the analysis. Therefore, we are simulating that a SM indicates the aggregated demand power data (lump load) from all the households that are connected to same phase in the multi-household property. This value is used by OpenDSS to calculate its respective voltage. 

\subsection{Predicting the voltage distribution}

Equation \ref{eq:predict} indicates the input to output correlation of the predictive model, with $f(.)$ is a 6-layer DLNN.

\begin{equation}
\hat{V}_{q}(t+\tau) = f(d_{q},H_q,t,I_N,X_m)
\label{eq:predict}
\end{equation}

\begin{equation}
X_m = \{x_1, x_2, x_c, ..., x_C\} 
\label{eq:xses}
\end{equation}

\begin{equation}
x_c = \{d_c, H_c, V_c, P_c\}
\label{eq:data}
\end{equation}

\noindent $\hat{V}_q$ is the predicted voltage for time $t+\text{30mins}$ for the queried CCP with the distance
 $d_{q}$ from source and the aggregated number of households $H_q$ between the source and $d_q$. $C$ is the number of CCP with SMs.
  
There are $N$ neurons in the input layer, followed by $N/2$ neurons in the $1^{st}$ hidden layer and $N/4$ neurons in the $2^{nd}$ to $4^{th}$ hidden layer. The output layer is a single neuron layer indicating the $\hat{V}_q$ value. 

The input is divided into four categories, beginning with (i) 3 neurons to indicate the queried CCP's $d_{q}$ and $H_q$, and $t$ for the time index of the measured SM data, (ii) 1 neuron for the total line impedance of the circuit $I_N$ to indicate its capacity and risk, (iii) $(C\times2\times5)$ or $(C\times1\times5)$ neurons are for the electricity measurements from $C$ number of available SMs and (iv) $(C\times2)$ neurons are to indicate the distance and loading corresponding to the $C$ number of SMs.
Therefore, the input layer consists of $N=4+(C\times2\times5)+(C\times2)$ or $N=4+(C\times1\times5)+(C\times2)$ number of neurons, depending if the demand power data are used as part of the input.

\subsubsection{Total Line Impedance, $I_N$}

The total line impedance of the circuit $I_N$ is calculated by first transforming the LV circuit into its equivalent electronic circuit, with each cable segment in the circuit appearing as a resistor with the impedance magnitude $Z = \sqrt{R^2+X^2}$. $Z$ is calculated using the cables’ resistance $Rm^{-1}$ and reactance $Xm^{-1}$ values provided by the cable manufacturer and the cable segment length $m$ reported in the Geographical Information System (GIS), which stores the network topologies. Because each cable has an impedance value $Z$, $I_N$ is then calculated using Th\'{e}venin's  equivalent circuit theorem.

An LV circuit is typically a 3-phase circuit with the customers assumed to be equally distributed across the 3 phases. However, when calculating $I_N$, all cables are assumed to be a single-phase cable and the customers are therefore all connected on to the same one-phase. Because of this, there will only be one $I_N$ per circuit, instead of 3, one for each of the phases. This single value will indicate the worst-case capacity for the LV circuit, representing the worst case in-balance situation when all customers are connected to a single common phase. 
This assumption is chosen because customers' phase data are often unavailable information.


\subsubsection{Electricity measurements and its respective loading}

In our analysis, a SM $c$ at a CCP with the distance $d_c$ from source provides the measurement data $x_c$ (\ref{eq:data}).
$x_c$ consists of the voltage magnitude $V_c$ and the aggregated active power $P_c$ at times $(t)$, $(t-\text{30mins})$, $(t-\text{1day})$, $(t-\text{30mins}-\text{1day})$ and $(t+\text{30mins}-\text{1day})$ for all the households that are connected to a specific phase at the CCP with SM $c$. $x_c$ also includes the distance from source $d_c$ and the aggregated number of households $H_c$ between the source and $d_c$. These two values are to indicate the loading which resulted in the voltage drop for time $t$ at location $d_c$. These values are useful when $P_c$ are not included as part of the input.

\begin{figure*}[t!]
	\centering
	\begin{subfigure}[b]{0.48\textwidth}
		\includegraphics[width=1\textwidth]{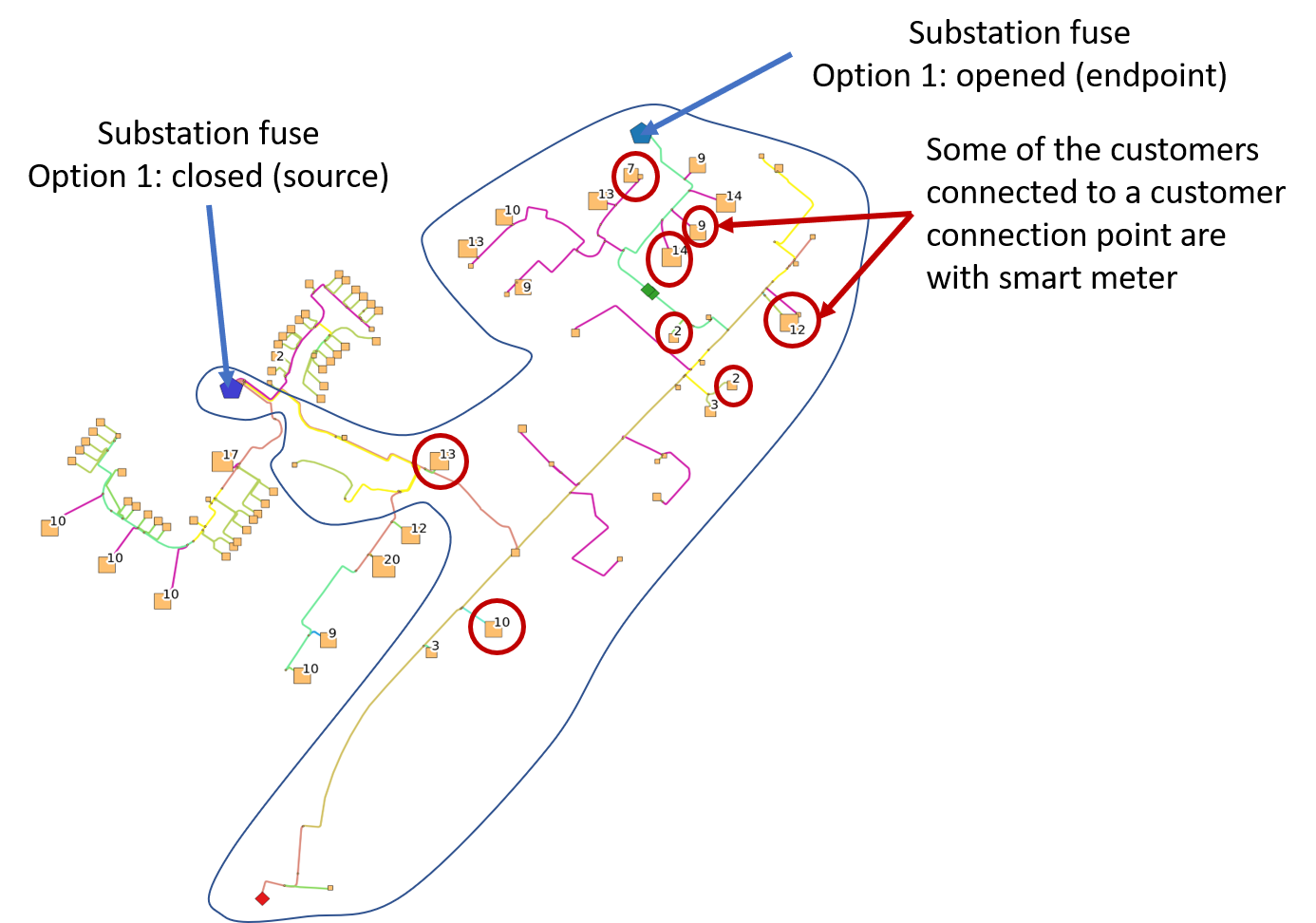}
		\caption{10 CCP}
		\label{fig:10ccp}
	\end{subfigure}
	\begin{subfigure}[b]{0.48\textwidth}
	\includegraphics[width=1\textwidth]{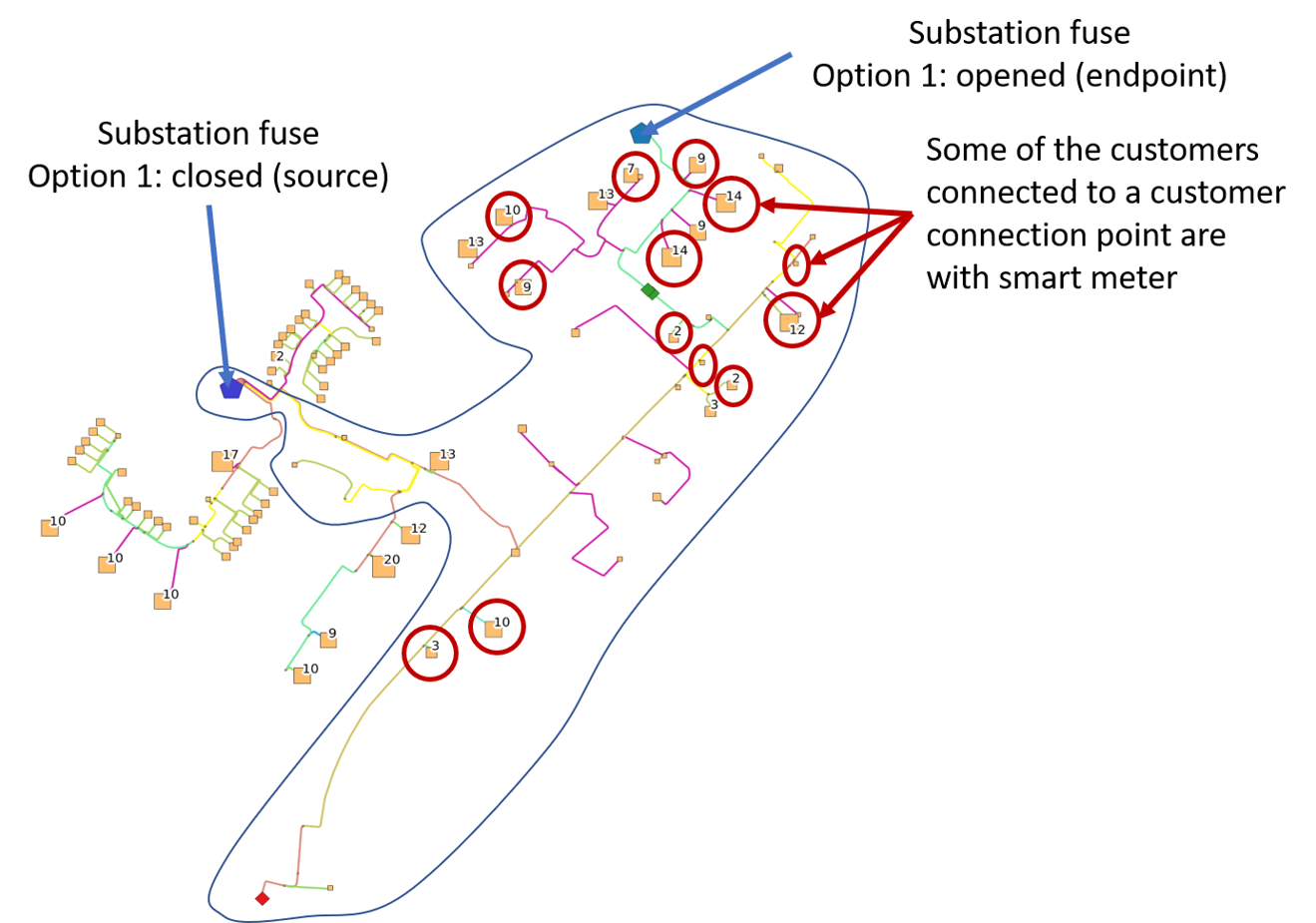}
	\caption{15 CCP}
	\label{fig:15ccp}
\end{subfigure}
	\caption{The customer connection points (CCPs) on the LV circuit with smart meter (SM) indicated by the red circles.}
	\label{fig:voltage}
\end{figure*}

\subsection{Deep Learning Neural Network (DLNN)}
DLNN has shown to be competitive for feature extraction and time-series analysis. For our analysis, the DLNN performs:
\begin{itemize}
	\item {\em Feature extraction:} to identify the correlation between the voltages provided by the SMs, their distances to source, and their approximated loading indicated by the aggregated number of households at the CCP and the aggregated demand power value if provided (\ref{eq:xses})--(\ref{eq:data}).  By identifying the correlation, the voltage for CCP without a SM can be approximated.
	\item {\em Time series analysis:} to identify how the voltage distribution changes over time. 
\end{itemize}

The DLNN was developed and trained using TensorFlow and keras libraries \cite{tensorflow}.
The activation function  used for all neurons is the Scaled Exponential Linear Unit. The DLNN is trained using the Adam optimiser with early stopping.



\subsection{Training the predictive model $f(.)$}
One-month demand profile data was simulated. 
The data from the first week for $C$ CCPs are used to train the DLNN. The data from the same $C$ CCPs from the following week are used for validation.
Only the $C$ CCPs are used to train and validate the DLNN, to simulate the lack of SM coverage. The $C$ CCPs are also randomly selected, to simulate the lack of control to the data availability from the SMs.
As indicated in Sec. \ref{sec:simulate}, the SM for single household CCP is similar to reality. However, for multi-households property, this will vary, whereby  any of the 3 single phase connection to the property are randomly selected to be that with a SM $c$ (\ref{eq:xses})--(\ref{eq:data}), $c \in C$.

\section{Results}
\label{results}

\begin{figure}[!t]
	\centering
	\includegraphics[width=0.48\textwidth]{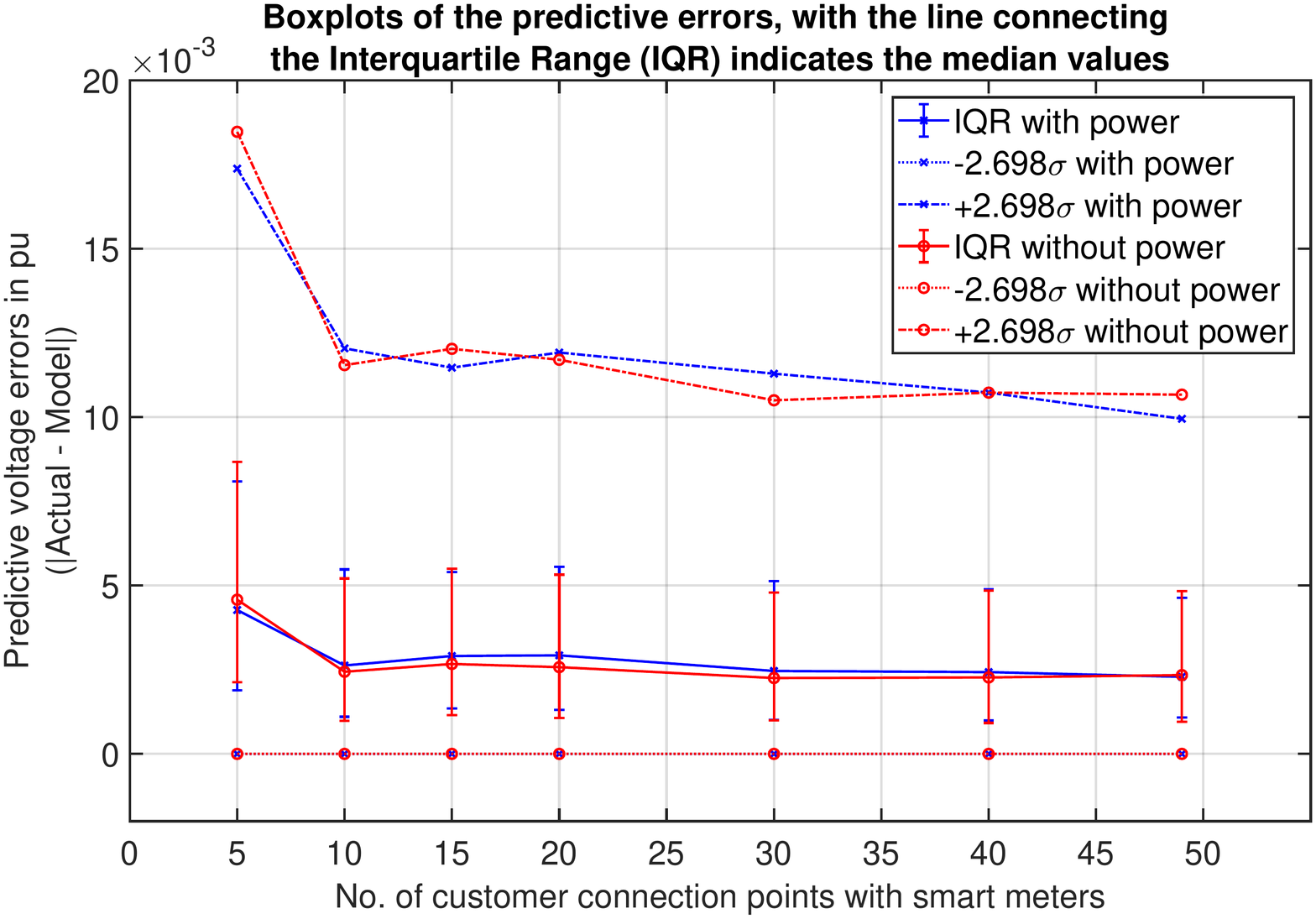}
	\caption{Predictive voltage error distributions for the varying no. of CCPs with SM for the LV circuit shown in Fig. \ref{fig:voltage}.}
	\label{fig:distr_single}
\end{figure}

Figure \ref{fig:distr_single} shows the predictive error distributions when the predictive model was created for varying $C$ number of CCPs with SM for the LV circuit shown in Fig. \ref{fig:voltage}. Figure \ref{fig:voltage} shows the location when there are $C=10$ and $C=15$ CCPs simulated with SMs.
The predictive error is calculated from all CCP in the circuit, with or without SM. Figure \ref{fig:distr_single}  aims to overlay two sets of distributions in the same figure, one when demand power data was included as part of the inputs to the DLNN (plots in blue) and one without (in red). To enable this, error bars are used to indicate the 
range between the $1^{st}$ and $3^{rd}$ quartiles or the interquartile range (IQR) of the distributions. The marker within the error bar shows the median value of the distribution. The line connecting the markers within the error bars connects the median values of the presented distributions. The figure also shows the $\pm2.698\sigma$ for each distribution.

\begin{figure}[!t]
	\centering
	\includegraphics[width=0.49\textwidth]{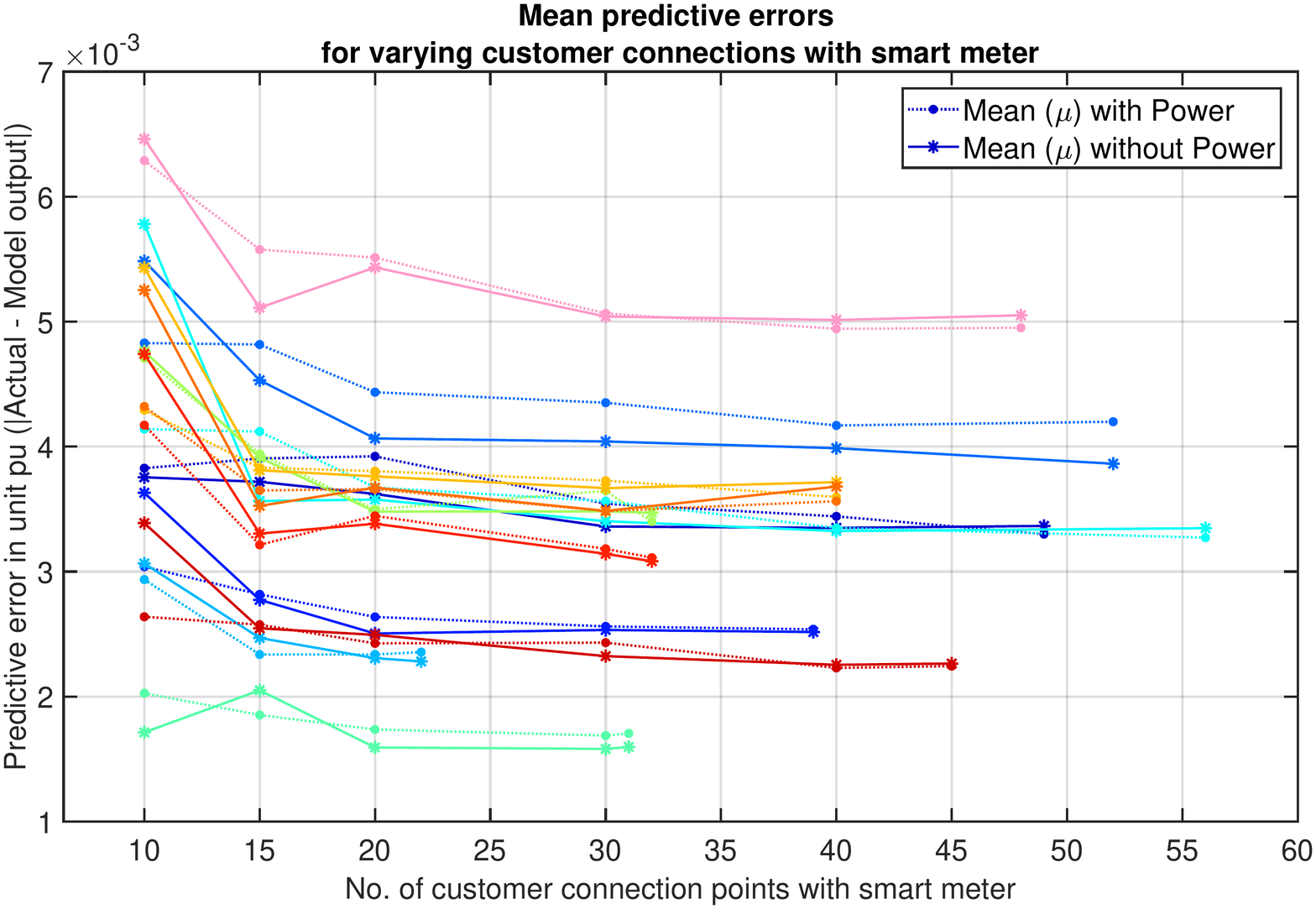}
	\caption{Mean predictive error for varying no. of CCPs with SM for 12 LV circuits, each represented by a specific colour.}
	\label{fig:distr_single2}
\end{figure}

The figure shows error distributions are similar after 10 CCPs, with or without the inclusion of the demand power data as part of the inputs. This indicates that data from key locations within a circuit are sufficient to enable effective prediction of voltage distribution, which are indicated in Fig \ref{fig:voltage}.
Based on this figure, the key locations identified are (i) first customer on the circuit; and (ii) the first and last customers on each branch. 
The median distance between SMs for the two $C$s are for $C=10$ CCPs is 52m and $C=15$ CCPs is 6m.

DLNNs are also created for 12 other circuits, each for the varying number of $C$ CCPs selected with SM, ending with 100\% SM coverage. Their mean predictive errors are displayed in Fig. \ref{fig:distr_single2}. Each circuit is represented by a specific colour, with the dot marker indicating when the inputs to the DLNN are with the demand power data and asterisk are without (\ref{eq:data}). 
The results show small dissimilarities between the mean predictive errors. 
There are also small dissimilarities shown when $C$ is greater than  a specific value ($C\geq$15).
This indicates the advantage of the proposed approach, which shows that if there are sufficient smart meter data available to provide observability to the LV circuit, individual and high granularity personal demand data are not required for DNO to predict the voltage distribution. This is because the DLNN is able to approximate the voltage drop at the queried CCP based on the key CCPs measurements and loading in the circuit. 

\begin{figure}[!t]
	\centering
	\includegraphics[width=0.48\textwidth]{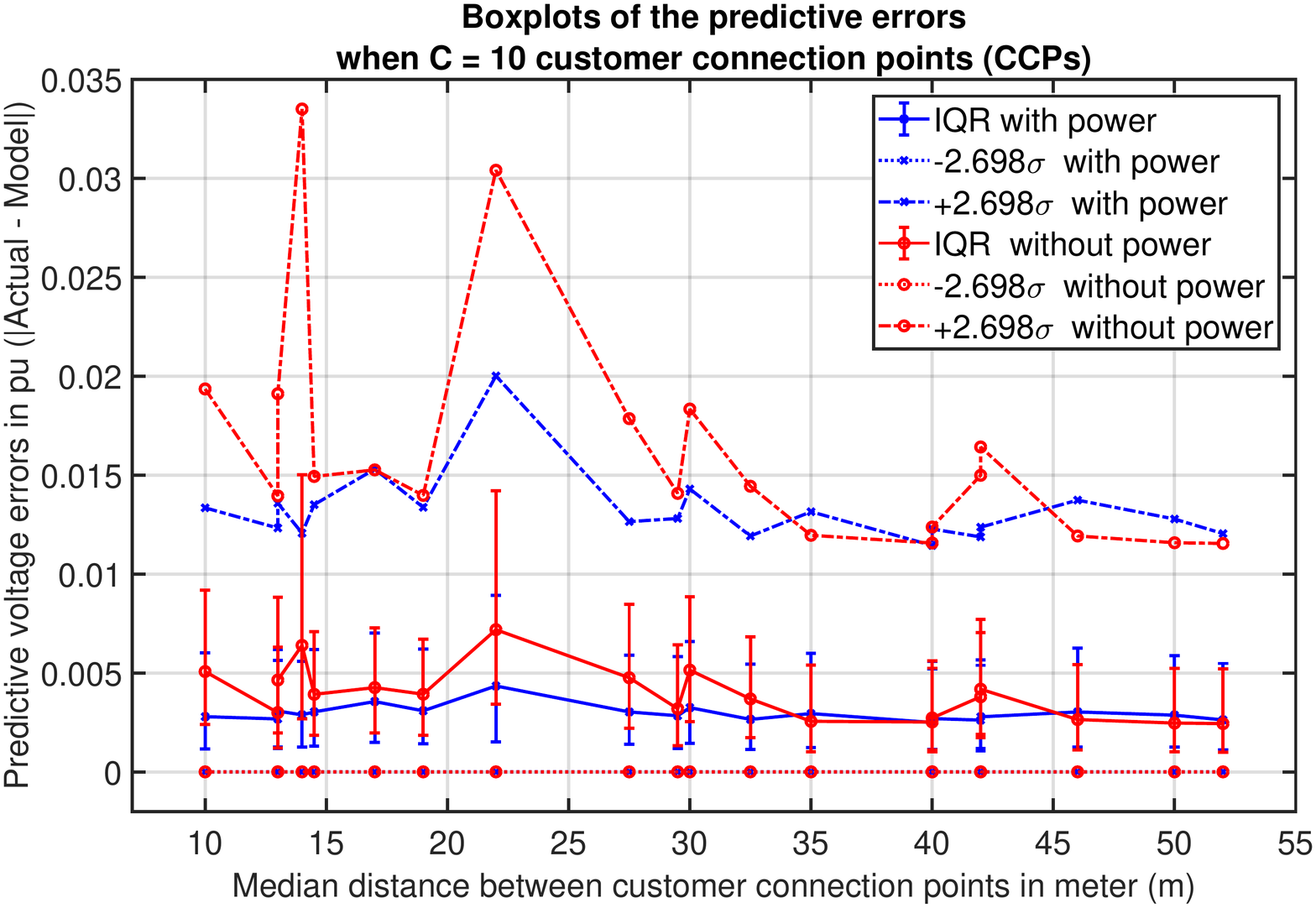}
	\caption{Predictive voltage error distributions when the locations of the $C=10$ CCPs with SM (Fig. \ref{fig:voltage}) were varied. The varying location is indicated by the median distance between the CCPs with SM.}
	\label{fig:vary_10}
\end{figure}

Identification of the key locations for SMs is key as this impact on the distribution of the predictive errors. Figure \ref{fig:vary_10} shows the impact of varying the location of the SMs, when $C=10$ CCPs with SM. Error bars shown in Fig. \ref{fig:vary_10} provide similar IQR and median value indication with that shown in Fig. \ref{fig:distr_single}, with the plots in blue indicating  when demand power data is included as part of the inputs to the DLNN and those in red are without. For $C=10$ CCPs, the larger the median distance between SMs in an LV circuit, the lower the likelihood that the SMs are clustered together within a specific area, resulting in limited SM coverage for the circuit and higher predictive errors. Those with high $+2.698\sigma$, 
the first CCP in the circuit is without a SM.
This indicates that if one wish to utilise the predictive model specifically without the use of demand power data, the SMs should ideally be  well spaced and at the key location within the circuit. This is as shown in Fig. \ref{fig:voltage}.

\section{Conclusions}
\label{concludes}

How energy is distributed across a Low-Voltage (LV) circuit is projected to change, and the previously passive `fit-and-forget' approach to network management will be inefficient to ensure its effective operations. An  adaptive approach is required that includes the prediction of risk to the circuits. Most methods described in literature require full observability of the networks. This premise is unrealistic in operation given the low-likelihood of full smart meter (SM) coverage for all the LV networks. This, together with privacy requirements have resulted in the low uptake of many distribution system state estimation methods for LV network analysis. To address this issue, we proposed the use of Deep Learning Neural Network (DLNN)  to predict how voltage is distributed across an LV circuit despite the partial SM coverage. The results show the applicability of the DLNN to create the predictive model, and that with SM data at key locations within the LV circuit is sufficient for effective prediction without requiring high granularity demand power data.




\bibliographystyle{IEEEtran}
\bibliography{bibliographies.bib}

\end{document}